\documentclass[10pt,twocolumn]{article} 
\usepackage{simpleConference}
\usepackage{amsmath,graphicx}
\usepackage{xcolor}
\usepackage{amssymb}
\usepackage{algorithm}
\usepackage{algpseudocode}
\usepackage{stackengine}
\usepackage{tabularx,booktabs}
\usepackage{graphicx}
\usepackage{multirow}
\usepackage{hyperref}

\usepackage{xcolor}

\usepackage[nomargin,inline,draft]{fixme}
\fxsetup{theme=color,mode=multiuser}
\FXRegisterAuthor{pf}{apf}{\color{red}pascal}

\DeclareMathOperator*{\argmin}{argmin}
\DeclareMathOperator*{\argmax}{argmax}
\makeatletter
\usepackage{forloop}
\newcounter{ct}
\newcommand{\markdent}[1]{\forloop{ct}{0}{\value{ct} < #1}{\hspace{-0.05cm}\hspace{\algorithmicindent}}}
\newcommand{\markcomment}[1]{\Statex\markdent{#1}}
\newcommand{\StateIndent}[1]{\State\markdent{#1}}

\usepackage{balance}
\begin{document}

\title{Targeted Adversarial Attacks against Neural Machine Translation}

\author{
Sahar Sadrizadeh$^1$, AmirHossein Dabiri Aghdam$^2$\thanks{\llap{}{The work has been done during an internship at LTS4, EPFL.}}, Ljiljana Dolamic$^3$, Pascal Frossard$^1$ \\\scalebox{0.9}{$^1${\textit{EPFL}}, Lausanne, Switzerland \enspace $^2$\textit{University of Tehran},Tehran, Iran
\enspace $^3$\textit{Armasuisse S+T}, Thun, Switzerland }
}

\maketitle
\thispagestyle{empty}

\begin{abstract}
Neural Machine Translation (NMT) systems are used in various applications. However, it has been shown that they are vulnerable to very small perturbations of their inputs, known as adversarial attacks. In this paper, we propose a new targeted adversarial attack against NMT models. In particular, our goal is to insert a predefined target keyword into the translation of the adversarial sentence while maintaining similarity between the original sentence and the perturbed one in the source domain. To this aim, we propose an optimization problem, including an adversarial loss term and a similarity term. We use gradient projection in the embedding space to craft an adversarial sentence. Experimental results show that our attack outperforms Seq2Sick, the other targeted adversarial attack against NMT models, in terms of success rate and decrease in translation quality. Our attack succeeds in inserting a keyword into the translation for more than 75\% of sentences while similarity with the original sentence stays preserved\footnote{The source code of our attack can be found at \href{https://github.com/sssadrizadeh/NMT-targeted-attack}{https://github .com/sssadrizadeh/NMT-targeted-attack}}.
\end{abstract}


\section{Introduction}

In spite of the impressive performance of Deep Neural Networks (DNNs) in different fields, from computer vision to Natural Language Processing, these models are shown to  be vulnerable to adversarial attacks, i.e., small perturbations of the input data \cite{szegedy2014intriguing}. In recent years many works have been proposed to evaluate the robustness of DNN models, and design methods to make them more robust to the perturbations to their inputs \cite{moosavi2016deepfool, madry2018towards, ortiz2021optimism,sadrizadeh2022block,sadrizadeh2023transfool}.

Neural Machine Translation (NMT) models, which take an input sentence and automatically generate its translation, have reached impressive performance by using DNN models such as transformers \cite{bahdanau2015neural, vaswani2017attention}. Due to their performance, NMT models are widely used in different applications. However, faulty outputs of such models may pose serious threats, especially in security-important applications.  Adversarial attacks against NMT models have been studied in the recent literature. First, the works in \cite{belinkov2018synthetic,ebrahimi2018adversarial}  show that character-level NMT models are vulnerable to character manipulations such as substitution or permutation. Moreover, the authors of  \cite{ebrahimi2018adversarial} also investigated  targeted attacks against NMT models, by which  they try to push/remove words from the translation. However, character manipulations and typos can be easily detected. Hence, most adversarial attacks against NLP and NMT systems use a word replacement strategy instead. Due to the discrete nature of textual data and difficulty of characterizing imperceptible  perturbation in text, 
most of the adversarial attacks against NLP models are based on heuristics. In particular, they first select some words in the input sentence and replace them with similar words to change the output of the target model. Michel et al. \cite{zhang2021crafting} and Zhang et al. \cite{michel2019evaluation}, propose untargeted  attacks to reduce the translation quality of NMT models. They first rank the words in the sentence based on their effect on the output, and substitute the important ones with similar words in the embedding space.  Cheng et al. \cite{cheng2019robust,cheng2020advaug} also propose an untargeted attack based on word replacement, in which they select  random words in the sentence and replace them with suitable substitutions by using a language model.  Wallace et al. \cite{wallace2020imitation}, propose a similar approach for generating adversarial examples with different objectives (i.e., universal untargeted attack), in which they find a universal trigger that is likely to fool the target model if added to the beginning of the sentence.   Since these methods are based on heuristics and word replacement, they may not achieve optimal attack performance. Therefore, Cheng et al. \cite{cheng2020seq2sick} propose a targeted adversarial attack based on optimization 
to fool the target model into generating translations that do not overlap with the original translation, or to push some words into the translation. They propose a hinge-like loss term to insert a keyword into translation, and a group lasso regularization to constrain the perturbation. However, the similarity term is in the embedding space of the target NMT model, which may not preserve semantic similarity.

In this paper, we study the problem of white-box targeted attacks against NMT models. We aim to generate adversarial examples that force the target model to insert some keywords in the translation. This type of attack is more dangerous but harder to build for the adversary than an untargeted attack, where the goal is merely to reduce the translation quality. 
Our method is based on optimization and gradient projection. 
We propose to use the embedding representation of a Language Model (LM), which captures the semantics of the tokens to compute the similarity between tokens. This is useful in building a similarity term in our optimization objective function. The objective function also includes an adversarial loss term to ensure that a keyword is present in the translation. To increase the chance of crafting a successful adversarial example, we do not fix a priori the position of the target word. Instead, we let the algorithm find the best position. 

We finally solve the proposed optimization problem iteratively as follows. Due to the discrete nature of the textual data,  we first perform a gradient descent step in the embedding space (which is continuous), then we perform a projection step to find the closest valid token. We use the embedding representation of a language model to find the closest token for the projection step. We consider two types of target keywords: predefined keywords, and $n
^{th}$-most likely class attack \cite{ebrahimi2018adversarial}, which chooses the $n
^{th}$-most probable token as the keyword. 
Experimental results demonstrate that our attack outperforms Seq2Sick, the only other white-box targeted attack against NMT models, in terms of success rate and decrease in translation quality. Moreover, our attack is more similarity-preserving for more challenging attack scenarios, e.g., predefined malicious keywords. 

The rest of this paper is organized as follows. We present the problem formulation and our optimization problem in \ref{sec:optimization}. Our attack algorithm is presented in Section  \ref{sec:method}. Section \ref{sec:experiment} provides the experimental setup and the results in comparison to Seq2Sick. Finally, the paper is concluded in Section \ref{sec:conclusion}.
\label{sec:intro}

\begin{figure*}[tb]
	\centering
    \scalebox{0.87}{
	\includegraphics[page=1,width=\linewidth, trim={4.7cm 3.5cm 3.2cm 4.7cm},clip]{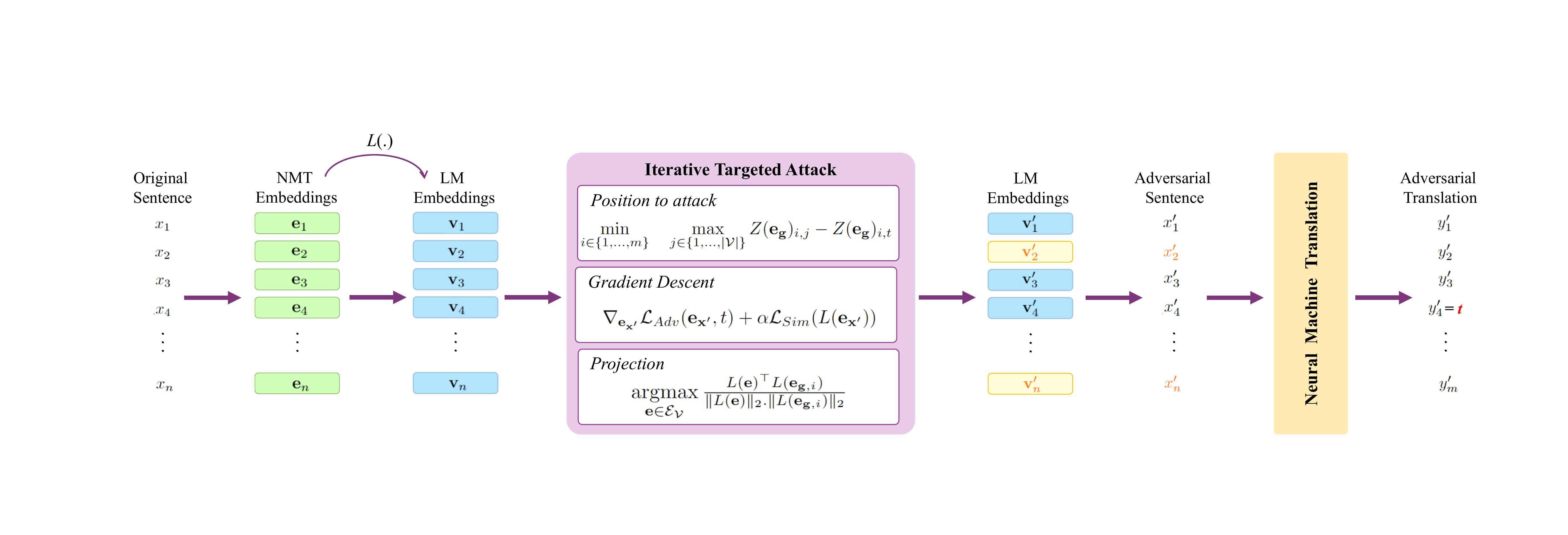}
	}\vspace{-5pt}
	\caption{Block diagram of the proposed method.}
	\label{fig:blockdiagram}
	\vspace{-10pt}
\end{figure*}

\section{Targeted Attack Problem}

In this section, we propose an optimization problem to craft adversarial examples against NMT models in targeted  attacks. 

NMT models get a sequence of tokens (i.e., words, subwords, or characters) in a source language, and convert them to a sequence of tokens in a target language. Let $f: \mathcal{X}\rightarrow \mathcal{Y}$ be the NMT model, which maps the input sentence $\mathbf{x}$ to its correct translation  $\mathbf{y}$. Every sentence is split into a set of tokens from the vocabulary set $\mathcal{V}$, that is,  the input sentence $\mathbf{x} = x_1x_2...x_n$ is a sequence of $n$ tokens, and its translation $\mathbf{y} = y_1y_2...y_m$ is a sequence of $m$ tokens. We aim to craft an adversarial sentence $\mathbf{x'}$, which we assume also has $n$ tokens,  by perturbing only a few tokens of $\mathbf{x}$ while inserting a predefined keyword $t$ into its translation:  $t \in \mathbf{y'}= f(\mathbf{x'})$. 

NMT models act like a classifier for all the tokens in the translation, and the classes are the tokens in the vocabulary set $\mathcal{V}$.  Therefore, for each word in the translated sentence, the output of the classifier (after the softmax function) is a probability vector over the vocabulary set. The inputs of the softmax function are called \textit{logits} and for the input sentence, we denote them by $Z(\mathbf{x}) =  \mathbf{z}_1\mathbf{z}_2...\mathbf{z}_n$, where $\mathbf{z}_i \in \mathbb{R}^{|\mathcal{V}|}$ is the logit vector for the $i^{th}$ token. In order to force the target translation model to include the target keyword $t$ in the translation $\mathbf{y'} = f(\mathbf{x'})$, we need to minimize the cross-entropy loss for one of the tokens, e.g., at position $k$, in the translation:
\begin{equation}
    \mathcal{L}_{Adv}(\mathbf{{x'}},t) =  \mathcal{L}_{f,k}(\mathbf{{x'}},t), \label{eq.ladv}
\end{equation}
where ${L}_{f,k}$ is the cross-entropy loss for the $k^{th}$ token in the translation when the target class is $t$. However, in order to increase the chance of generating a successful adversarial sentence with the keyword $t$, it is reasonable to choose the position of the target keyword in the translation $k$ based on the logits. In other words, we choose the position $k$ such that it is the easiest to insert the keyword  compared to other positions in the translation. This simply means that we are looking for the position $k$, where the difference of the logits between the most probable token and target keyword $t$ is minimized:
\begin{equation} \label{pos}
    k =  \min_{i\in\{1,...,m\}} \quad \max_{j \in \{1,...,|\mathcal{V}|\}}\mathbf{z}_{i,j} - \mathbf{z}_{i,t}.
\end{equation}

However, if we only minimize the cross-entropy loss of Eq. \eqref{eq.ladv} to find an adversarial example, we may end up with a very different sentence than the original sentence, while we would like to keep perturbations undetectable. Hence, we add a similarity constraint to the optimization problem. To define this similarity constraint, we propose to use a language model since it has been shown that language models capture the semantics of the tokens \cite{tenney2019you}. Generally, the tokens, which are in a discrete space, are  transformed into a continuous vector, known as the embedding vector. Let  $\mathbf{v}_i$ and  $\mathbf{v}'_i$ denote the embedding vectors by the LM for the $i^{th}$ tokens of the input and adversarial sentences, respectively. We can find the distance between two sentences by computing the average of cosine distances between the corresponding LM embedding vectors:
\begin{equation}\label{eq.lsim}
    \mathcal{L}_{Sim}(\mathbf{{v'}}) = \frac{1}{n}\sum_{i=1}^{n} 1-\frac{\mathbf{v}_i^\intercal\mathbf{v}'_i}{\|\mathbf{v}_i\|_2.\|{\mathbf{v}'_i}\|_2}.
\end{equation}
Cosine distance is zero for the same tokens and it has higher values for  unrelated tokens. The minimization of $\mathcal{L}_{Sim}$, which equals to the $\ell_1$ norm (summation)  of cosine distances for all tokens, ensures that only a few tokens are perturbed.

Finally, we propose the following optimization problem to generate an adversarial example with small perturbations in the original sentence, such that the translation contains a predefined target keyword $t$:
\begin{equation} \label{optimization}
 \mathbf{x'} = \argmin_{\substack{\mathbf{x}'_i\in \mathcal{V}}} \mathcal{L}_{Adv}(\mathbf{{x'}},t) + \alpha \mathcal{L}_{Sim}(\mathbf{{v'}}), 
\end{equation}
where $\alpha$ is the hyperparameter that controls the importance of the similarity constraint in the optimization, and $\mathcal{L}_{Adv}$ and $\mathcal{L}_{Sim}$ are defined in \eqref{eq.ladv} and \eqref{eq.lsim}, respectively.
\label{sec:optimization}

\section{Attack Algorithm}

In this section, we present our attack algorithm to solve the proposed optimization problem.
The optimization problem of Eq. \eqref{optimization} is discrete since we are dealing with the tokens of the adversarial sentence that are in a discrete space. We propose to solve the optimization in the embedding space of the target NMT model, which is continuous. 
The block-diagram of our attack is depicted in Figure \ref{fig:blockdiagram}. We first decompose the input sentence into a sequence of tokens, convert them to continuous embedding vectors, and then find  a transformation between the embedding spaces of the target NMT model and a language model.  Afterwards, we use gradient projection to solve the optimization problem of Eq. \eqref{optimization}. In particular, we perform a gradient descent step of our optimization problem in the embedding space. Since the resulting embedding vectors may not necessarily correspond to a meaningful and valid token, we then project the resultant embedding vectors to the nearest embedding vectors that correspond to valid tokens. 

\begin{figure}[!t]
\begin{small}
\scalebox{0.82}{
\begin{minipage}{0.52\textwidth}
\begin{algorithm}[H]
\caption{Targeted Attack against NMT Models.}
\label{alg}
\begin{algorithmic}[1]
\State \textbf{Input}:
\markcomment{1} $f(.)$: Target NMT model, $\mathcal{V}$: Vocabulary set
\markcomment{1} $\mathbf{x}$ : Tokenized input sentence, $lr$: Learning rate
\markcomment{1} $\alpha$: hyperparameter, $L$: Linear function
\markcomment{1} $K$: Maximum No. of iterations, $t$: Target keyword 
\State \textbf{Output}:
\markcomment{1} $\mathbf{x'}$: Generated adversarial example
\markcomment{1} \textbf{initialization:}
\StateIndent{1} $\textbf{b}\leftarrow\text{empty}$,  $itr \leftarrow 0$
\StateIndent{1} $\forall i \in \{1,...n\} \quad \mathbf{e}_{\mathbf{g},i}\leftarrow \mathbf{e}_i$ 

\While {$itr \le K$}
\State $itr \leftarrow itr + 1$
\markcomment{1} \textit{Best position to attack:}
\StateIndent{0}$k \leftarrow  \min\limits_{i\in\{1,...,m\}} \quad \max\limits_{j \in \{1,...,|\mathcal{V}|\}}Z(\mathbf{e_g})_{i,j} - Z(\mathbf{e_g})_{i,t}$
\vspace{2pt}
\markcomment{1} \textit{Gradient descent:}
\StateIndent{0} $\mathbf{e_g} \leftarrow \mathbf{e}_{\mathbf{g}} - lr. \nabla_{\mathbf{e_{x'}}} \mathcal{L}_{Adv}(\mathbf{e_{x'}},t) + \alpha \mathcal{L}_{Sim}(L(\mathbf{e_{x'}}))$
\markcomment{1} \textit{Projection:} 
\StateIndent{0} \textbf{for} {$i \in \{1,...,n\}$} \textbf{do}
\StateIndent{1} $\mathbf{e}_{\mathbf{p},i} \leftarrow  \argmax\limits_{\mathbf{e} \in \mathcal{E_V}} \frac{L(\mathbf{e})^\top L(\mathbf{e}_{\mathbf{g},i})}{\|L(\mathbf{e})\|_2.\|L(\mathbf{e}_{\mathbf{g},i})\|_2}$
\StateIndent{0} \textbf{end for}
\StateIndent{0} \textbf{if} {$\mathbf{e_p}$ not in \textbf{b}} \textbf{then}
\StateIndent{1} add $\mathbf{e_p}$ to \textbf{b}
\StateIndent{1} $\mathbf{e_g}\leftarrow \mathbf{e}_{\mathbf{p}}$
\StateIndent{0} \textbf{end if}
\If {$t \in f(\mathbf{e}_{\mathbf{p}})$}
\State\Return ${\mathbf{e_{x'}}} \leftarrow \mathbf{e_p}$ (adversarial example is found)
\EndIf

\EndWhile

\end{algorithmic}
\normalsize

\end{algorithm}
\end{minipage}
}\end{small}
\end{figure}

The pseudo-code of our attack algorithm is presented in Algorithm \ref{alg}. Let us denote the NMT embedding vectors of the adversarial sentence by $\mathbf{e_{x'}} = [ \mathbf{e}'_1,..., \mathbf{e}'_n]$, where $\mathbf{e}'_i$ is the NMT embedding representation  of the $i^{th}$ token of the adversarial sentence. We can find its LM embedding vector by a linear function $\mathbf{v}'_i = L(\mathbf{e}'_i)$\footnote{In order to find this transformation, we train a language model alongside the linear layer $L$ that gets the NMT embedding vectors in the input.}. To insert the target keyword into the translation, we first find the best position to attack. Instead of fixing the attack position at the beginning of the attack according to Eq. \eqref{pos}, we find the best position in each iteration of the algorithm (line 7). In other words, we consider the computed adversarial example after each iteration of the algorithm as the sentence we want to attack; and hence, find the best position again in every iteration. Afterwards, we update the embedding vectors of the adversarial sentence by moving in the opposite direction of the
gradients (line 9). Then we perform a projection step to $\mathcal{E_V}$, which is the discrete subspace of the embedding space containing the embedding vectors of  every token in the vocabulary set $\mathcal{V}$ (line 11). We use the embedding representation of the language model to find the closest similar token. We should note that we only perform the projection step if the resulting adversarial sentence is new, in order to prevent the algorithm from getting stuck in a loop of previously computed sentences (line 13). We perform the gradient descent and projection step iteratively until the translation contains the  keyword $t$ or a maximum number of iterations is reached. If the computed adversarial sentence after a maximum number of iterations does not contain the keyword $t$, we consider the attack to be unsuccessful.

\label{sec:method}

\section{Experimental Results}

In this section, we evaluate the performance of our attack against different translation models and translation tasks. We compare the results of our attack with that of Seq2Sick \cite{cheng2020seq2sick}. Seq2Sick is the only white-box targeted attack against NMT models, to the best of our knowledge.

For the target NMT model, we attack  HuggingFace \cite{wolf-etal-2020-transformers} implementations of Marian NMT models \cite{junczys-dowmunt-etal-2018-marian}. 
We conduct experiments on the test set of the wmt14 dataset \cite{bojar2014findings} for the English to French (En-Fr) and English to German (En-de) tasks. We report the results for 1000 randomly chosen sentences from each dataset. Some statistics of these datasets can be found in Table \ref{tab:dataset}. 

\setlength{\textfloatsep}{5pt}
\begin{table}[tbp]
	\centering
		\renewcommand{\arraystretch}{0.8}
	\setlength{\tabcolsep}{4pt}
	\caption{Some statistics of the evaluation datasets.}
	\label{tab:dataset}
	\scalebox{0.79}{
		\begin{tabular}[t]{@{} lcccc @{}}
			\toprule[1pt]
		    \multirow{1}{*}{\textbf{Dataset}}  &    \multirow{1}{*}{{\textbf{Avg. Length}}} &
		    \multirow{1}{*}{{\textbf{\#Test Samples}}} & 
		    \multirow{1}{*}{\textbf{BLEU score}} \\
		    
			
			\midrule[1pt]
			WMT14 (En-Fr)  &  \multirow{1}{*}{27} & \multirow{1}{*}{3003} & \multirow{1}{*}{39.88} \\ 
		
			\midrule
			WMT14 (En-De)  &  \multirow{1}{*}{26} & \multirow{1}{*}{3003} & \multirow{1}{*}{27.72} \\

			\bottomrule[1pt]
		\end{tabular}
	}
\end{table}

We use the Adam optimizer with a learning rate of $0.02$ to solve the optimization problem. We also set hyperparameter $\alpha\in \{10,4,1\}$. If the attack is not successful for a high value of $\alpha$, we perform the attack again with a lower value to make the attack more aggressive. Finally, in order to find the  linear function $L(.)$ that converts the NMT embeddings to LM embeddings, we finetune GPT-2 pretrained language model alongside the linear function $L(.)$ on WikiText-103 dataset with causal language modeling objective function.  

In order to evaluate our adversarial attack, we consider two types of keywords. In the first one, we assume that the target keyword is predefined. In this case, we consider the word \textit{war} in French (guerre) and German (krieg), which can be a malicious word. 
However, as opposed to classification task, in NMT models, we are dealing with $|\mathcal{V}|$ number of classes, which is at least tens of thousands. Therefore, it is harder for the adversary to find a successful adversarial example. The $n^{th}$-most likely class attack is defined in \cite{ebrahimi2018adversarial}, in which the target word is the $n^{th}$ most probable token at a fixed position of the translation. In our second evaluation, we consider this type of attack with the difference that we choose the position for the attack such that the $n^{th}$ likely token has the minimum \textit{logit} difference with the most probable token compared  to other positions in the translation. 

\begin{table}[tbp]
	\centering 
		\renewcommand{\arraystretch}{1}
	\setlength{\tabcolsep}{2.5pt}
	\caption{Performance of targeted attack for various keywords
	.}
	\label{tab:result}
	\scalebox{0.75}{
		\begin{tabular}[t]{@{} lcccccccc @{}}
			\toprule[1pt]
		    \multirow{2}{*}{\textbf{Keyword}}  &
		    \multirow{2}{*}{\textbf{Method}}  &
		    \multicolumn{3}{c}{\textbf{En-Fr}} &&   \multicolumn{3}{c}{\textbf{En-de}} \\
			\cline{3-5}
			\cline{7-9}
			\rule{0pt}{2.5ex}    
			 & & \scalebox{0.95}{ASR$\uparrow$} & \scalebox{0.95}{RDBLEU$\uparrow$} &  \scalebox{0.95}{Sim.$\uparrow$}  && \scalebox{0.95}{ASR$\uparrow$} &  \scalebox{0.95}{RDBLEU$\uparrow$} & \scalebox{0.95}{Sim.$\uparrow$}   \\
			\midrule[1pt]
            \multirow{2}{*}{\textit{war}} & ours  & \textbf{99.29} &  0.20 & \textbf{0.83} && \textbf{83.84} & \textbf{0.29} & \textbf{0.77} \\ 
			
			 & Seq2Sick & 86.68 & \textbf{0.28} &  0.73 && 27.41 & 0.22 & 0.74 \\  
            \midrule[0.5pt]
			 \multirow{2}{*}{$2^{nd}$} & ours & \textbf{76.13} & \textbf{0.17} & 0.91 && \textbf{78.60} & \textbf{0.20} & 0.90 \\ 
			 & Seq2Sick & 69.26 & 0.10 & \textbf{0.92} && 68.66 & 0.13 & \textbf{0.92} \\ 
			
			\midrule[0.5pt]
			\multirow{2}{*}{$10^{th}$} & ours & \textbf{80.84} & \textbf{0.21}  & 0.87  && \textbf{75.25} & \textbf{0.26}  & \textbf{0.86} \\ 
			
			 & Seq2Sick & 63.93  & 0.17 & \textbf{0.88} && 61.71 & 0.20 &  0.86\\ 
            \midrule[0.5pt]
			 \multirow{2}{*}{$100^{th}$} & ours & \textbf{85.50} & \textbf{0.23} & \textbf{0.84}  && \textbf{80.83} & \textbf{0.27} & \textbf{0.83} \\ 
			 & Seq2Sick & 61.87 & 0.21 & 0.83 && 62.58 & 0.22 & 0.81  \\

			\bottomrule[1pt]
		\end{tabular}
	}
\end{table}

\begin{table*}[!t]
	\centering
		\renewcommand{\arraystretch}{.88}

	\setlength{\tabcolsep}{3pt}
	\caption{An adversarial example against En-Fr Marian NMT  for $2^{nd}$-most likely attack (target keyword: programme).}
	\scalebox{0.73}{
		\begin{tabular}[t]{@{} l| >{\parfillskip=0pt}p{21.5cm} @{}}
			\toprule[1pt]
		    \textbf{Sentence}  
		    & \textbf{Text}\\
		
			\midrule[1pt]
			
		    \multirow{1}{*}{Org.} 
		    &  Mr Dutton called on Ms Plibersek to guarantee that not one dollar out of the rescue package would be spent on additional bureaucracy. \hfill\mbox{}  \\
			 
			\cline{2-2}
			\rule{0pt}{2.5ex} 
			
			
			
			\multirow{1}{*}{Org. Trans.} 
			& M. DUTTON demande à Mme Plibersek de garantir qu'aucun dollar du plan de sauvetage ne sera dépensé pour une bureaucratie supplémentaire. \hfill\mbox{} \\
			 
			\cline{1-2}
			\rule{0pt}{2.5ex}

			\multirow{1}{*}{Adv. Ours} 
			&  Mr Dutton called on Ms Plibersek to guarantee that not one dollar out of the rescue package would be spent on additional \textcolor{red}{\textbf{workforce}}. \hfill\mbox{} \\
			
			\cline{2-2}
			\rule{0pt}{2.5ex} 
			
			\multirow{1}{*}{Trans.} 
			& M. DUTTON demande à Mme Plibersek de garantir qu'aucun dollar du \textcolor{red}{\textbf{programme}} de sauvetage ne sera \textcolor{orange}{consacré à la main-d'œuvre} supplémentaire. \hfill\mbox{} \\
			 
			\cline{1-2}
			\rule{0pt}{2.5ex}

			\multirow{1}{*}{Adv. Seq2Sick} 
			&  Mr Dutton called on Ms Plibersek to guarantee that not one dollar out of the rescue \textcolor{red}{\textbf{Programs}} would be spent on additional bureaucracy. \hfill\mbox{} \\
			
			\cline{2-2}
			\rule{0pt}{2.5ex} 
			
			\multirow{1}{*}{Trans.} 
			& M. DUTTON demande à Mme Plibersek de garantir qu'aucun dollar \textcolor{orange}{des programmes} de sauvetage ne sera dépensé pour une bureaucratie supplémentaire. \hfill\mbox{}  \\

			\bottomrule[1pt]
			
		\end{tabular}
	}
	\label{tab:sample}
	
\end{table*}

We evaluate the performance in terms of different metrics: Attack Success Rate (ASR), semantic Similarity (Sim.) between the adversarial example and the original sentence computed by universal sentence encoder \cite{yang2020multilingual}, and Relative Decrease of the BLEU score (RDBLEU) \cite{post-2018-call}. These metrics are computed over the successful attacks and the results are reported in   Table \ref{tab:result}. 
Overall, our attack is able to insert the target word into the  translation for more than 75\% of the sentences in all cases. At the same time, it maintains similarity with the original sentence (more than 0.8) in almost all cases. As this Table shows, by increasing $n$ in $n^{th}$-most likely attack, the similarity of the adversarial examples with the original sentence decreases while the success rate increases. Moreover, our attack outperforms Seq2Sick in terms of success rate and relative decrease in BLEU score. Additionally, the ability of our attack in preserving similarity with the original sentence is competitive with Seq2Sick and ours is better especially when the attack is harder such as $n=100$ in $n^{th}$-most likely attack and when the target keyword is \textit{war}. This may be due to the fact that we use LM embedding vectors instead of NMT ones for similarity constraint.

\begin{figure}[tb]
	\centering
	\includegraphics[page=1,width=.9\linewidth, trim={0cm 0cm 0cm 0cm},clip]{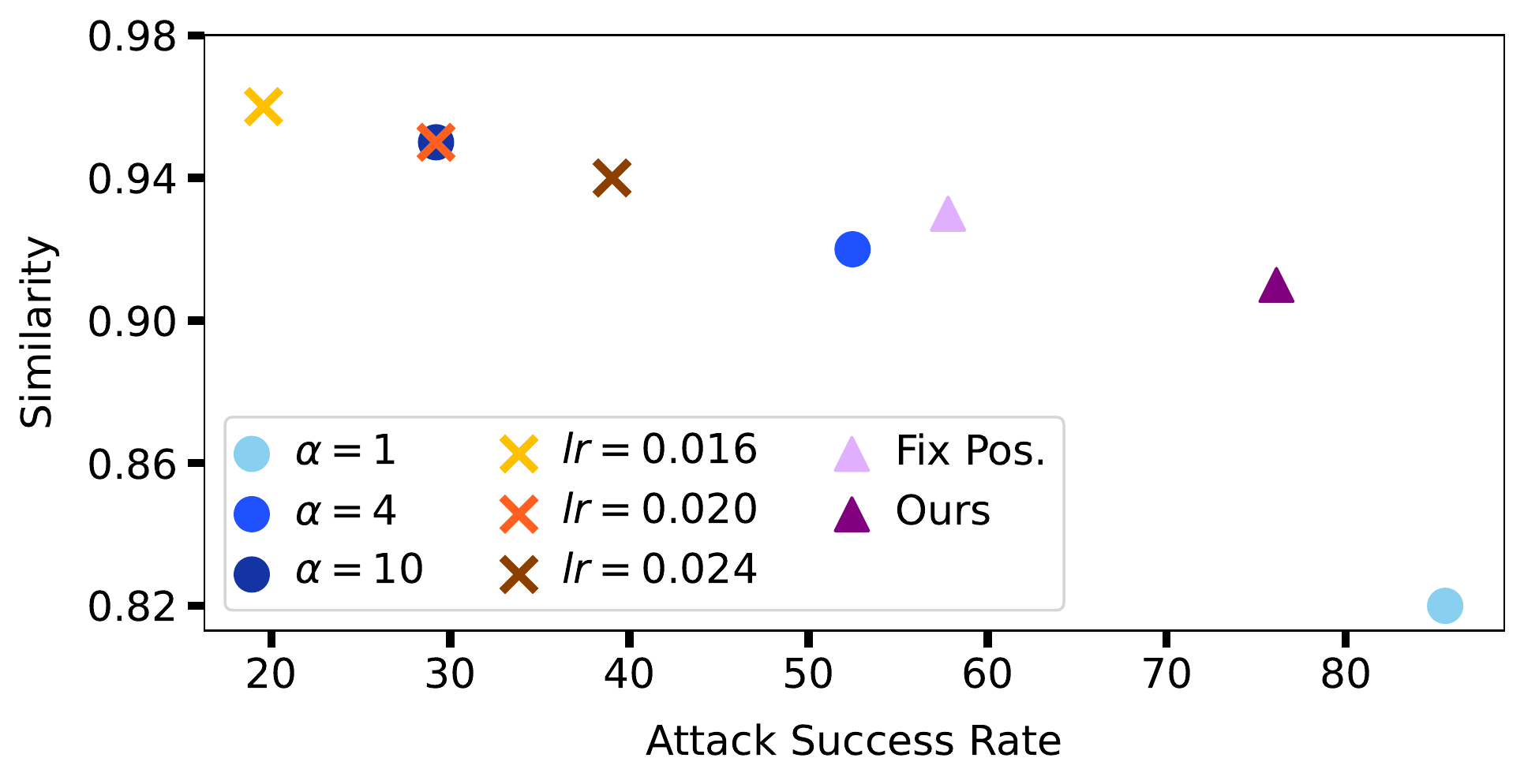}
 \caption{Affect of hyperparameters in $2^{nd}$-most likely attack.
	}
	\label{fig:abl}
\end{figure}

An adversarial example for the case of $2^{nd}$-most likely attack, with the the target keyword \textit{programme}, is presented in Table \ref{tab:sample}. As this example shows, the change made by our attack is more subtle and less related to the target keyword as opposed to the changes made by Seq2Sick. 
Figure \ref{fig:abl} shows the effect of different hyperparameters on our attack performance. By increasing the learning rate $lr$ and decreasing $\alpha$, the similarity constraint coefficient in our optimization,  the attack becomes more aggressive. Hence, the success rate increases while similarity with the original sentences decreases. Moreover, suppose we fix the position of the target keyword in the translation, as opposed to our strategy to find the best position in each iteration. In that case, the success rate highly drops, which shows the importance of this step of our attack algorithm. The run-time of our attack to insert \textit{guerre} is 8.2 seconds on a system equipped with an NVIDIA A100 GPU. However, Seq2Sick takes 38.4 seconds to craft an adversarial sentence, which is more time-consuming than our attack.

\label{sec:experiment}

\section{Conclusion}

In this paper, we proposed a new white-box targeted attack  against NMT models. First, we proposed an optimization problem to force the NMT model to insert a keyword into the translation and preserve similarity with the original sentence. Then, we introduced an iterative algorithm to solve the optimization and craft an adversarial example. Experimental results show that our attack is highly effective in different translation tasks and target keywords. We also compared our method with Seq2Sick, the only other white-box targeted attack against NMT models. Our attack outperforms Seq2Sick in terms of success rate and decrease in translation quality while they are both able to preserve similarity.

\label{sec:conclusion}

\small
\balance
\bibliographystyle{abbrv}
\bibliography{main}

\end{document}